\documentclass[11pt]{article}
\usepackage{smureport}
\let\cite\citep  

\usepackage{xcolor}
\usepackage{colortbl}
\usepackage{pifont}
\usepackage{multirow}
\usepackage{wrapfig}

\providecommand{\cmark}{}
\renewcommand{\cmark}{\textcolor{green!80!black}{\ding{51}}}
\providecommand{\xmark}{}
\renewcommand{\xmark}{\textcolor{red!55!black}{\ding{55}}}
\newcommand{\namark}{\textcolor{gray}{n/a}}
\definecolor{oursrow}{HTML}{EBF4FE}  
\definecolor{grouprow}{HTML}{EFEFEF}  

\title{ManimCode: Learning Visual Self-Correction for Educational Animation}

\author{Zhengyang Liang\textsuperscript{1}\quad Lizi Liao\textsuperscript{1}}

\smuaffiliations{\textsuperscript{1}Singapore Management University}
\smuinstitute{}  

\smudate{October 31, 2026}
\smuhomepage{https://liang-zhengyang.github.io/manimcode}  
\smucontact{\href{mailto:zhengyang.l.2025@phdcs.smu.edu.sg}{zhengyang.l.2025@phdcs.smu.edu.sg}}  
\smushorttitle{ManimCode}
\hypersetup{pdftitle={ManimCode: Learning Visual Self-Correction for Educational Animation},
  pdfauthor={Zhengyang Liang, Lizi Liao},pdfsubject={Educational animation generation}}

\newcommand{\taboneformat}{\scriptsize\setlength{\tabcolsep}{3pt}}  
\newcommand{\supplref}[1]{Appendix~\ref{#1}}

\makeatletter
\newcommand{\keepspace}[1]{\par\penalty-100\begingroup\setlength{\dimen@}{#1}\dimen@ii\pagegoal\advance\dimen@ii-\pagetotal
  \ifdim\dimen@>\dimen@ii\ifdim\dimen@ii>\z@\vfil\fi\break\fi\endgroup}
\makeatother

\begin{document}
\addtocontents{toc}{\protect\setcounter{tocdepth}{-1}}
\maketitle
\begin{smuabstract}
Manim supports precise, programmatic video generation, but a successfully rendered educational animation can still have visual defects, missing explanations, or incorrect content. Rendering does not directly identify the code changes needed to correct these visual failures, so feedback-driven revision is needed. Existing agentic frameworks use iterative rendering and revision but typically rely on large models, often proprietary. Existing Manim datasets primarily contain final programs and often lack quality verification and the trajectories of feedback and revisions needed to learn this refinement process.
We introduce ManimData and ManimCode, a data and training framework for feedback-driven educational animation generation. We construct ManimData through a render-guided pipeline that iteratively generates, executes, reviews, and revises Manim programs with retrieved API documentation. Its 24,034 render-verified, quality-filtered records preserve intermediate programs, execution errors, visual critiques, and corrective revisions. We use role-conditioned supervised fine-tuning on these trajectories to train ManimCode, a single 4B model, to write, review, and revise animation programs. To improve correction of errors that differ from those in the collected trajectories, we then train the Reviser with reinforcement learning on self-generated states, using rendering outcomes and reference-free quality feedback.
ManimCode achieves 96.6\% and 97.0\% render success on ManimData-test and ManimBench, compared with 52.6\% and 82.0\% for the untrained 4B model under the same inference setup, while also improving video quality scores from 2.48 to 2.94 and from 3.00 to 3.44. Integrating ManimCode into two long-form educational video generation systems increases completed videos from 19/117 to 107/117 and from 53/239 to 239/239, alongside improvements in educational effectiveness and visual quality.
\end{smuabstract}

\clearpage
\section{Introduction}
\label{sec:intro}

Educational animations help explain abstract STEM concepts by showing mathematical relationships, geometric constructions, and processes that unfold over time. Creating these animations by hand, however, requires both subject knowledge and programming expertise. Recent work uses language models to write programs for animation engines such as Manim, which render them into videos~\cite{pmlr-v306-chen26bx,ku-etal-2025-theoremexplainagent}. This approach offers precise control over formulas, text, and geometry, which are difficult to maintain with pixel-space video generation. A successful render can still produce a poor teaching video: labels may overlap, objects may fall outside the frame, or explanations may omit important steps. These defects are difficult to fix because the video does not directly identify the required code changes. Producing a usable animation therefore requires inspecting the output and revising the program based on the defects.

Agentic systems such as Code2Video~\cite{pmlr-v306-chen26bx}, TheoremExplainAgent~\cite{ku-etal-2025-theoremexplainagent}, and ManimAgent~\cite{jiang2026manimagent} use execution errors and, in some cases, visual feedback to revise generated programs. These systems typically rely on large models, often proprietary, and their refinement loops become much less reliable with smaller open alternatives. Specialized training is also limited by the available data. ManimBench~\cite{rammuni2026manim} contains render-verified programs but is relatively small, while community datasets and JanusCode-800K~\cite{sun2026januscoder} lack systematic rendering and quality checks. Of the 20,976 Hugging Face community programs we collected, roughly one-third fail to render as released. Even among the renderable Hugging Face programs, only 31\% pass our quality criterion after revision. These datasets retain final programs but discard the errors, feedback, and edits that led to them. ManimTrainer~\cite{silva2026manimtrainer} uses reference-based rewards during training and execution feedback at inference, but does not directly supervise models to interpret and correct visual defects. Existing data therefore offers little guidance on how a model should review and revise its own animations.

To make feedback-driven refinement learnable by small open models, we build \textbf{ManimData} through a render-guided pipeline that records how animation programs are corrected and improved. We collect Manim programs from public datasets and generate additional requests from Wikipedia STEM topics. A strong open multimodal model then writes or revises each program, executes it, and reviews the rendered frames. Execution errors, layout checks, and visual critiques guide subsequent revisions, with relevant API documentation retrieved from a knowledge base covering Manim and eight domain plugins. The pipeline keeps the best version and admits a record only when it meets the quality criterion. The resulting 24,034 render-verified, quality-filtered records contain the final program and video, intermediate versions, errors, rendered frames, reviews, and revisions, preserving the feedback and corrective actions missing from existing Manim corpora.

We use ManimData to train \textbf{ManimCode}, enabling a single 4B model to write, review, and revise animation programs. We first organize the recorded trajectories into role-conditioned supervised fine-tuning (SFT) tasks. The \emph{Writer} generates programs from requests and retrieved API documentation, the \emph{Reviewer} assesses rendered frames and identifies visual defects, and the \emph{Reviser} corrects execution failures and improves programs based on visual critiques. Training on final programs and intermediate correction steps improves initial generation and feedback-driven revision. Because the trajectories come from a stronger model, they may not fully capture errors encountered by the smaller model. We address this gap by training the Reviser with reinforcement learning (RL) on self-generated repair and critique states, using rendering outcomes and reference-free quality feedback as rewards. At inference, a single 4B model runs the complete render--review--revise loop without the larger model.

We evaluate ManimCode on two scene-level animation benchmarks and integrate it as the scene generator in two long-form systems, Code2Video and TheoremExplainAgent. It achieves 96.6\% and 97.0\% render success on ManimData-test and ManimBench, compared with 52.6\% and 82.0\% for the untrained 4B backbone under the same inference setup. With the planning and rendering budgets held constant, completed videos increase from 19/117 to 107/117 and from 53/239 to 239/239, respectively, alongside improvements in learning-gain and visual-quality metrics.

Our contributions are:
\begin{itemize}
    \item We introduce \textbf{ManimData}, a dataset of 24,034
    render-verified, quality-filtered animations. Alongside final
    programs, its trajectories preserve execution errors, visual
    critiques, and corrective revisions, directly supervising models
    to identify and fix generation failures.

    \item We propose \textbf{ManimCode}, a framework that trains
    a single 4B model to run the complete render--review--revise process.
    Role-conditioned trajectory SFT teaches it to generate, critique,
    and correct animation programs; reinforcement learning on
    self-generated failures improves correction of errors encountered
    at inference.

    \item ManimCode achieves 96.6\% and 97.0\% render success
    on two scene-level benchmarks. When integrated into two existing
    long-form educational video generation systems, it increases
    video completion.
\end{itemize}

\section{Related Work}
\label{sec:related}

\subsection{Feedback-Driven Code Generation}
Language models can improve programs at inference time by critiquing and revising their own outputs~\cite{madaan2023selfrefine,shinn2023reflexion} or by reading execution results~\cite{chen2024selfdebug,ding2024cycle}, although under matched sampling budgets such self-repair often gains little over resampling~\cite{olausson2024selfrepair}. Later work trains these abilities directly. RLEF~\cite{gehring2024rlef} and SCoRe~\cite{kumar2025score} optimize multi-turn correction with reinforcement learning from execution feedback or final-answer correctness, while CriticGPT~\cite{mcaleese2024criticgpt} and CTRL~\cite{xie2025ctrl} train critics that are rewarded for the bugs they catch or the fixes they enable. When a program's output is visual, rendering supplies the feedback. Training corpora for chart, plotting, and user-interface code are filtered or repaired through execution~\cite{zhao2025chartcoder,ni2025viscoder,wu2024uicoder}, and reinforcement learning methods derive rewards for rendered outputs from their similarity to reference images~\cite{rodriguez2025rlrf,chen2025rrvf}, from a multimodal critic~\cite{li2025relook}, or from verifiable layout metrics~\cite{pan2026aeslides}. Closest to our setting, WebLoop~\cite{meng2026webloop} trains one policy to generate, critique, and refine web pages, and UI2Code$^N$~\cite{yang2025ui2coden} and VisRefiner~\cite{deng2026visrefiner} train models to refine UI code from rendered screenshots. Most of these methods target static images or web pages, and most train only the generator or only the critic. In our setting, programs render videos whose defects unfold over time, usually with no reference to compare against, and one model learns to critique the rendered frames as well as to write and revise the program.

\subsection{Code-Based Educational Animation}
Code-based generation is now a common route to precise educational animation. Agentic systems write Manim scenes with large, mostly proprietary language models~\cite{samarth2025manimator}. Code2Video~\cite{pmlr-v306-chen26bx} and TheoremExplainAgent~\cite{ku-etal-2025-theoremexplainagent} plan long-form lectures and repair each scene's code from render errors over several renders, and ManimAgent~\cite{jiang2026manimagent} guides the repairs with visual review of the rendered frames. OmniManim~\cite{li2026seebeforeyoucode} adds a learned layout prior, SGA~\cite{lopez2026sga} checks scene geometry symbolically before rendering, and multi-agent systems such as SciEducator~\cite{xu2026scieducator} produce multimodal educational content from scientific videos. Benchmarks score generated Manim code on executability, request alignment, and visual similarity~\cite{rammuni2026manim,oli2026manibench}.

Training-based work is sparse. JanusCoder~\cite{sun2026januscoder} includes Manim programs in a large corpus of visual code. ManimTrainer~\cite{silva2026manimtrainer} follows fine-tuning with reinforcement learning that rewards similarity to reference programs and videos, and at inference it re-prompts the model with render errors and retrieved documentation. SimuScene~\cite{wang2026simuscene} rewards physics animations with a vision-language model that answers verification questions about the rendered video. Each of these trains a model to write a complete program, and feedback enters only as a training reward or as error messages at inference. ManimData instead records how programs were reviewed and revised, so our model learns these steps as well.

\section{Data}
\label{sec:manimdata}

\begin{figure*}[t]
    \centering
    \includegraphics[width=\linewidth]{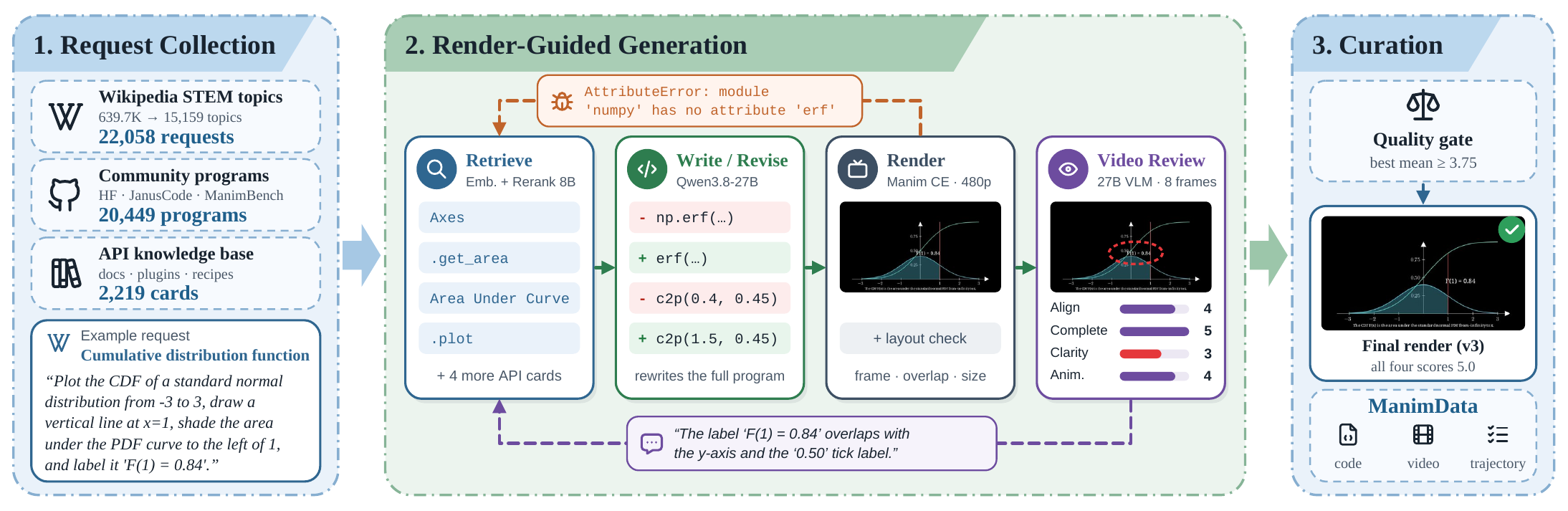}
    \caption{\textbf{ManimData construction pipeline.} (1) Request collection: Wikipedia STEM requests, community programs, and the API knowledge base. (2) Render-guided generation: Qwen3.8-27B writes or revises the program from retrieved API cards, and each version is rendered, checked for layout defects, and reviewed on eight frames; an error or a critique (a crash and an overlapping label) starts the next revision. (3) Curation: an input whose best version reaches a mean review score of at least 3.75 is admitted with its code, video, and trajectory.}
    \label{fig:manimdata-pipeline}
\end{figure*}

ManimData pairs each request with a render-verified program, its video, and the trajectory of errors, critiques, and revisions that produced the program. It draws on two sources: requests derived from Wikipedia STEM topics and public community programs (Sec.~\ref{sec:manimdata-sources}). A single render-guided pipeline writes, renders, checks, reviews, and revises the programs from both sources with retrieved API documentation (Sec.~\ref{sec:manimdata-curation}, Fig.~\ref{fig:manimdata-pipeline}). Table~\ref{tab:manimdata-comparison} compares ManimData with existing Manim datasets and benchmarks.

\begin{table*}[t]
    \centering
    \caption{Comparison with existing Manim datasets and benchmarks.
    \namark{} marks a property that does not apply to the evaluation-only long-form benchmarks.}
    \label{tab:manimdata-comparison}
    \taboneformat
    \renewcommand{\arraystretch}{1.2}
    \begin{tabular*}{\textwidth}{
        @{\extracolsep{\fill}}l r c l c c c@{}
    }
        \toprule
            \multirow{2}{*}{Dataset}
            & \multirow{2}{*}{Size}
            & \multirow{2}{*}{Purpose}
            & \multirow{2}{*}{Target Video}
            & Render & Quality & Revision \\
            & & & & Verified & Filtered & Trajectories \\
        \midrule
        ManimBench \cite{rammuni2026manim}
            & 416 & Train/Eval & Scene-level Animation
            & \cmark & \xmark & \xmark \\
        JanusCode-800K (Manim) \cite{sun2026januscoder}
            & 8,316 & Train & Scene-level Animation
            & \xmark & \xmark & \xmark \\
        HF Community (7 datasets)
            & 20,976 & Train & Scene-level Animation
            & \xmark & \xmark & \xmark \\
        \midrule
        MMMC (Code2Video) \cite{pmlr-v306-chen26bx}
            & 117 & Eval & Long-form Explanation
            & \namark & \namark & \xmark \\
        TheoremExplainBench \cite{ku-etal-2025-theoremexplainagent}
            & 239 & Eval & Long-form Explanation
            & \namark & \namark & \xmark \\
        \midrule
        \textbf{ManimData (Ours)}
            & 24,034 & Train & Scene-level Animation
            & \cmark & \cmark & \cmark \\
        \bottomrule
    \end{tabular*}
\end{table*}

\subsection{Data Sources and API Knowledge Base}
\label{sec:manimdata-sources}

\noindent\textbf{Wikipedia requests.}
We derive requests from 639,700 English Wikipedia articles. A coarse-to-fine filter with Qwen3.5-4B~\cite{qwen35} and Qwen3.6-27B~\cite{qwen36_27b} keeps 15,159 topics whose central ideas can be shown graphically. Qwen3.8-27B~\cite{qwen38} finds directly drawable content in 12,516 of these topics, assigns each of them one of 30 visualization types, and writes up to two open-ended requests per topic specifying what to draw and emphasize. Excluding requests that depend on realistic biological or medical illustration (e.g., cells) leaves 22,058 requests across 11,150 topics (\supplref{app:sources}).

\noindent\textbf{Community programs.}
We also collect programs from seven Hugging Face datasets (HF Community), the Manim split of JanusCode-800K~\cite{sun2026januscoder}, and the ManimBench training split~\cite{rammuni2026manim}. From HF Community we keep only the programs that render as released. After deduplication, 20,449 inputs remain, and each starts from its released code.

\keepspace{2.5\baselineskip}
\noindent\textbf{API knowledge base.}
We compile 2,219 API cards (\supplref{app:kb}) so the Writer and Reviser can look up APIs for our renderer, ManimCE v0.19.1 with eight domain plugins. The cards come from the official documentation, descriptions of plugin objects written by Qwen3.8-27B, capability recipes that map topic keywords to plugins, and the best-practice rules and example scenes of \texttt{manim\_skill}. Cards are retrieved with Qwen3-Embedding-8B and reranked with Qwen3-Reranker-8B~\cite{qwen3embedding}.

\subsection{Render-Guided Generation and Filtering}
\label{sec:manimdata-curation}

Every input enters the render--review--revise loop (Fig.~\ref{fig:manimdata-pipeline}), in which Qwen3.8-27B plays the \emph{Writer}, the \emph{Reviewer}, and the \emph{Reviser} within a budget of $N=10$ renders. Let $x$ be the request and $c$ the current program. For a Wikipedia request, the Writer writes retrieval queries for $x$ and drafts $c$ from the retrieved API cards $\mathcal{K}$; a community input starts from its released code. The renderer $\mathcal{R}$ either produces a video $v=\mathcal{R}(c)$ or stops with an execution error $e$. If $c$ renders, a layout checker reads object positions at animation pauses and reports geometric defects as layout diagnostics $\ell$: objects outside the frame, overlapping text, and labels too small to read.

\begin{wrapfigure}{r}{0.42\linewidth}
    \vspace{-12pt}
    \centering
    \includegraphics[width=\linewidth]{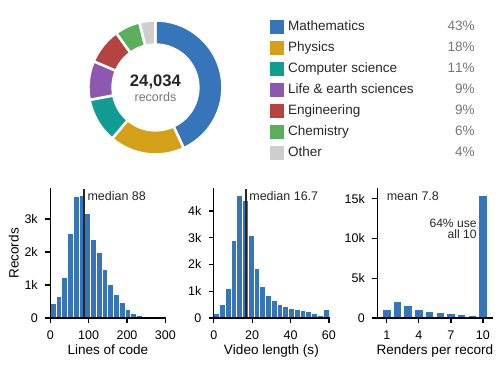}
    \caption{ManimData statistics. Top: domain shares of the 24,034 records. Bottom: per-record distributions of final program length, video length, and number of renders.}
    \label{fig:manimdata-analysis}
    \vspace{-10pt}
\end{wrapfigure}
The Reviewer then sees $x$, $c$, and eight frames sampled from $v$. It scores request alignment, content completeness, visual clarity, and animation quality from 1 to 5 and writes a critique of specific defects, including non-geometric ones such as missing content; the scores and critique form the review $f$. The Reviser rewrites the full program: it repairs a crash from $e$ or revises a rendered program from $f$ and $\ell$, with cards $\mathcal{K}$ retrieved anew for each revision.

We track the best candidate $c^*$ based on the review scores, retaining
a new version only if it improves the mean score without substantially
reducing request alignment or content completeness. We retain records whose best candidate achieves
a mean score of at least 3.75, storing the final program, video, and
complete refinement trajectory in ManimData. The loop stops when a high-quality result is confirmed
by a second review or the rendering budget is exhausted
(\supplref{app:loop}).

\subsection{Dataset Statistics}
\label{sec:manimdata-statistics}

ManimData contains 24,034 render-verified, quality-filtered
animation records from Wikipedia requests and community
programs. It covers diverse STEM topics, with mathematics,
physics, and computer science accounting for 72\% of the
records (Figure~\ref{fig:manimdata-analysis}). The median
program contains 88 lines of code and produces a 16.7-second
video. Most records undergo multiple refinement rounds,
preserving the errors, critiques, and corrections encountered
during generation. These trajectories yield 305,452
role-conditioned training examples for ManimCode.

\section{Method}
\label{sec:method}

ManimCode learns from both the final programs and the intermediate corrections in ManimData's render--review--revise trajectories. Role-conditioned trajectory supervision trains a 4B model to write programs, assess rendered outputs, and correct execution and visual defects. Because a stronger model collected these trajectories, we further train the Reviser with reinforcement learning on failures generated by the supervised model. At inference, the model iteratively refines educational animations within a fixed rendering budget.

\begin{figure*}[t]
    \centering
    \includegraphics[width=\linewidth]{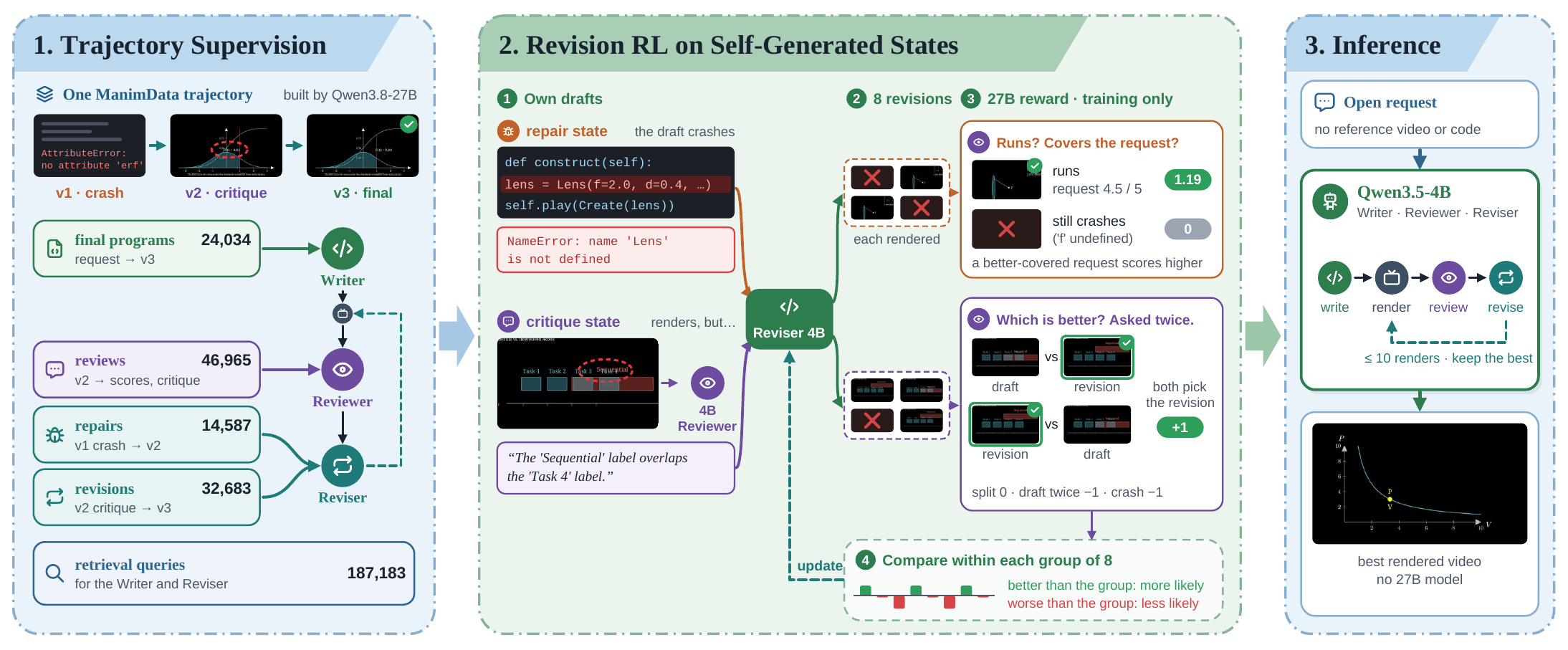}
    \caption{\textbf{Overview of ManimCode.} (1) Final programs and intermediate steps of ManimData trajectories supervise five tasks of one Qwen3.5-4B model. (2) Reinforcement learning further trains the Reviser on repair and critique states that the supervised model produces itself: the Reviser samples eight revisions per state, Qwen3.8-27B scores each rendered revision (request coverage for repair, two order-swapped comparisons with the draft for critique), and revisions that score above the rest of their group become more likely (Sec.~\ref{sec:method-rl}). The two states shown are real training states; their revisions and scores are illustrative. (3) At inference, our model runs the render--review--revise loop. Qwen3.8-27B is used only to build the data and to compute training rewards.}
    \label{fig:manimcode-overview}
\end{figure*}
\subsection{Role-Conditioned Trajectory Supervision}
\label{sec:method-sft}

We use each trajectory to supervise the model's writing, reviewing, and revision steps. Every example is a triple $(\rho,z,y)$: role $\rho$, context $z$, and target $y$. The \emph{Writer} receives the request and retrieved API cards and targets the trajectory's final program $c^*$, so it learns from the refinements made during construction. The \emph{Reviewer} receives the request, current program, and sampled frames; its targets are the four recorded scores and the critique. For the \emph{Reviser}, we extract two kinds of successful transitions: (i) \emph{execution repairs}, revisions that turn a crashed program into one that renders, and (ii) \emph{quality revisions}, critique-driven revisions for which the mean review score exceeds that of the program being revised. The Reviser targets the complete revised program given the request, program being revised, error or critique, layout diagnostics when available, and API cards. We also supervise retrieval queries based on the request and, when available, the error or critique, so the model can retrieve API cards at inference.

We use the same role instruction and context layout at training and inference. The SFT mixture $\mathcal{D}_{\mathrm{SFT}}$ contains 24,034 writing, 14,587 execution-repair, 32,683 quality-revision, 46,965 reviewing, and 187,183 retrieval-query examples. We train one Qwen3.5-4B model on this mixture by supervised fine-tuning (SFT), minimizing
\begin{equation}
    \mathcal{L}_{\mathrm{SFT}}(\theta)
    =-\mathbb{E}_{(\rho,z,y)\sim\mathcal{D}_{\mathrm{SFT}}}
    \log \pi_{\theta}(y\mid\rho,z),
    \label{eq:method-sft}
\end{equation}
where $\pi_\theta$ is the model with trainable parameters $\theta$.

\subsection{Revision RL on Self-Generated States}
\label{sec:method-rl}

The Reviser receives the model's own mistakes as input, so we train it with reinforcement learning on states generated by the supervised model. Rewards combine render outcomes with a reward model's judgments against the request or the input program. This setup requires no reference program or target video.

\paragraph{Self-generated training states.}
For educational requests disjoint from the evaluation sets, the supervised model retrieves API cards $\mathcal{K}$, writes a first draft, and renders it. If the draft fails to render, we define a \emph{repair state} $z^{\mathrm{rep}}=(x,c,e,\mathcal{K})$. If it renders, the same model reviews it. If the review does not meet the early-stopping criterion, its review $f$ and layout diagnostics $\ell$ form a \emph{critique state} $z^{\mathrm{crit}}=(x,c,f,\ell,\mathcal{K})$. Both states follow the Reviser's inference input format.

\paragraph{Rewards.}
For each state $z$, the Reviser samples $G=8$ candidate revisions $\{c'_i\}_{i=1}^{G}$, and we render each candidate. Near-duplicate revisions receive zero reward to discourage trivial changes. For repair states, a candidate that fails to render receives zero reward, while a successful render is rewarded based on its request alignment and content completeness:
\begin{equation}
    r_{\mathrm{rep}}(c') = 1 + \frac{q(c')-3}{8},
    \quad
    q(c')=\frac{q_{\mathrm{align}}+q_{\mathrm{complete}}}{2},
    \label{eq:method-repair-reward}
\end{equation}
where $q_{\mathrm{align}},q_{\mathrm{complete}}\in[1,5]$
are scores assigned by Qwen3.8-27B based on the request,
program, and rendered frames.
For critique states, where the original program already renders, we use pairwise preferences to assess whether a revision improves upon the original. A revision that fails to render receives $-1$, while successful revisions are compared with the original in both presentation orders to reduce position bias~\cite{zheng2023judging}. Let $b_1,b_2\in\{\mathrm{new},\mathrm{input}\}$ denote the two preferences. The reward is
\begin{equation}
 r_{\mathrm{crit}}(c')=\begin{cases}
  +1, & b_1=b_2=\mathrm{new},\\
  -1, & b_1=b_2=\mathrm{input},\\
  0, & \text{otherwise}.
 \end{cases}
 \label{eq:method-visual-reward}
\end{equation}
We optimize the Reviser using leave-one-out policy optimization (RLOO)~\cite{ahmadian2024rloo} with a clipped policy-gradient objective and a KL penalty
toward the supervised policy. Further training details are provided in \supplref{app:training}.

\paragraph{Inference.}
At inference, ManimCode follows the same render--review--revise loop used during data construction, with a single 4B backbone handling all three roles. Each request starts from a newly generated program, and the loop retains the best successfully rendered candidate within a budget.

\section{Experiments}
\label{sec:experiments}

We evaluate ManimCode with three research questions.
\textbf{RQ1:} How does ManimCode perform compared with existing general-purpose and Manim-specialized models on scene-level animation generation?
\textbf{RQ2:} Can ManimCode improve long-form educational video generation when integrated into existing systems?
\textbf{RQ3:} How do ManimData, trajectory supervision, and Reviser RL contribute to generation performance? We further analyze representative revision examples and examine the agreement between automatic quality assessments and human preferences.

\subsection{Experimental Setup}
\label{sec:exp-setup}

\begin{table*}[t]
    \centering
    \caption{
    Scene-level animation generation on ManimData-test and ManimBench.
    VQ is averaged over successfully rendered videos.
    $^{\dagger}$71 of the 100 ManimBench requests appear in
    JanusCoder's training data.
    }
    \label{tab:exp-scene-results}
    \scriptsize
    \setlength{\tabcolsep}{4pt}
    \setlength{\aboverulesep}{0pt}  
    \setlength{\belowrulesep}{0pt}
    \renewcommand{\arraystretch}{1.25}
    \newcommand{\grouprow}[1]{\rowcolor{grouprow}\multicolumn{7}{c}{\emph{#1}} \\}
    \begin{tabularx}{\textwidth}{>{\hsize=2.5\hsize\raggedright\arraybackslash}X
        >{\hsize=0.75\hsize\centering\arraybackslash}X
        *{5}{>{\hsize=0.75\hsize\raggedleft\arraybackslash}X}}
        \toprule
        & & \multicolumn{2}{c}{ManimData-test} & \multicolumn{3}{c}{ManimBench} \\
        \cmidrule(lr){3-4}\cmidrule(l){5-7}
        Model & Size & RS$\uparrow$ & VQ$\uparrow$ & RS$\uparrow$ & VS$\uparrow$ & VQ$\uparrow$ \\
        \midrule
        \grouprow{Large general-purpose models}
        GPT-4o & -- & 60.2 & 2.87 & 79.0 & 71.0 & 3.34 \\
        GLM-5.3 Flash & MoE & 57.2 & \textbf{3.18} & 83.0 & 73.7 & \textbf{3.51} \\
        DeepSeek-V4 Flash & MoE & 62.4 & 3.05 & 81.0 & 72.9 & 3.33 \\
        \grouprow{Small general-purpose models}
        Qwen3.5-4B & 4B & 7.0 & 2.34 & 31.0 & 27.6 & 3.01 \\
        Qwen3.5-9B & 9B & 15.4 & 2.60 & 46.0 & 40.4 & 3.24 \\
        Qwen3.8-27B & 27B & 39.8 & 3.09 & 59.0 & 52.6 & 3.37 \\
        \grouprow{Manim-specialized models}
        JanusCoder-8B$^{\dagger}$ & 8B & 67.0 & 2.55 & 71.0 & 64.8 & 3.21 \\
        ManimTrainer (Seed-Coder-8B) & 8B & 49.8 & 2.41 & 67.0 & 59.9 & 3.10 \\
        Qwen3.5-4B + our loop & 4B & 52.6 & 2.48 & 82.0 & 71.7 & 3.00 \\
        ManimTrainer (Seed-Coder-8B) + RITL-DOC & 8B & 56.6 & 2.38 & 73.0 & 65.0 & 3.05 \\
        \grouprow{Our trained model}
        \rowcolor{oursrow} Ours (SFT, single pass) & 4B & 64.4 & 2.82 & 81.0 & 69.7 & 3.29 \\
        \rowcolor{oursrow} Ours (SFT, loop) & 4B & 88.6 & 2.89 & 92.0 & 79.0 & 3.45 \\
        \rowcolor{oursrow} \textbf{Ours (SFT + RL, loop)} & 4B & \textbf{96.6} & 2.94 & \textbf{97.0} & 82.9 & 3.44 \\
        \bottomrule
    \end{tabularx}
\end{table*}

\paragraph{Benchmarks.}
We evaluate scene generation on two test sets. \emph{ManimData-test} consists of 500 requests selected from the ManimData request pool, 219 Wikipedia requests and 281 community requests, all excluded from training (\supplref{app:sources}).
\emph{ManimBench}~\cite{rammuni2026manim} has 100 test requests with reference programs, and we follow its protocol. Of the 100 references, 85 nearly duplicate ManimCE documentation examples that our knowledge base contains, so in all our ManimBench runs the retriever excludes those examples. For long-form generation we use \emph{MMMC}, the 117 lecture topics of Code2Video~\cite{pmlr-v306-chen26bx}, and \emph{TheoremExplainBench} (TEB), the 239 distinct theorems of TheoremExplainAgent~\cite{ku-etal-2025-theoremexplainagent}.

\paragraph{Baselines.}
The general-purpose baselines served through APIs are GPT-4o~\cite{openai2024gpt4o} and two recent mixture-of-experts models, GLM-5.3 Flash and DeepSeek-V4 Flash, which we query through OpenRouter\footnote{\url{https://openrouter.ai/}}. The open general-purpose baselines that we run locally are Qwen3.5-4B (our base model), Qwen3.5-9B, and Qwen3.8-27B. The Manim-specialized baselines are JanusCoder-8B~\cite{sun2026januscoder}, whose training data contains most ManimBench test requests ($^{\dagger}$ in Table~\ref{tab:exp-scene-results}), and the released Seed-Coder-8B checkpoint of ManimTrainer~\cite{silva2026manimtrainer}. Two further baselines repair their programs at inference. ManimTrainer + RITL-DOC runs ManimTrainer's own repair loop with documentation retrieval. Qwen3.5-4B + our loop puts the untrained model in every role of our loop, with the same retrieval and budget as our system.

\paragraph{Metrics and judge.}
For scene-level evaluation, we report render success (RS), visual similarity (VS), and video quality (VQ). RS measures the percentage of successfully rendered programs. On ManimBench, VS measures similarity to reference videos using the geometric mean of SSIM and CLIP similarities, following \citet{silva2026manimtrainer}. VQ is evaluated by Gemini 3.8 Flash\footnote{\url{https://deepmind.google/models/model-cards/gemini-3-8-flash/}} on a 1--5 scale, averaging content quality (request alignment, completeness, and correctness) and presentation quality (visual clarity and animation quality). VQ is computed only over successfully rendered videos. For long-form generation, we report TeachQuiz learning gain and aesthetics on MMMC and visual quality on TEB. We use the same judge for both benchmarks and exclude narration-based dimensions from TEB. Further evaluation details are provided in \supplref{app:eval}.

\subsection{Main Results}
\label{sec:exp-main}

\paragraph{Scene-level generation.}
Table~\ref{tab:exp-scene-results} compares ManimCode with general-purpose and Manim-specialized models. With a 4B generator, ManimCode achieves 96.6\% and 97.0\% render success on ManimData-test and ManimBench, respectively, outperforming the evaluated general-purpose models (at most 62.4\% and 83.0\%) and released Manim-specialized systems (at most 67.0\% and 73.0\%).
ManimCode's VQ scores are comparable to GPT-4o and approach those of recent MoE models, particularly on ManimBench.

\paragraph{Contribution of each component.}
We examine the contributions of the refinement loop, trajectory SFT, and Reviser RL using the 4B variants in Table~\ref{tab:exp-scene-results}. Applying our loop to the untrained Qwen3.5-4B raises render success from 7.0\% to 52.6\% on ManimData-test and from 31.0\% to 82.0\% on ManimBench, with only modest changes in video quality. Under the same loop and retrieval setup, trajectory SFT further improves render success to 88.6\% and 92.0\%, while also increasing VQ from 2.48 to 2.89 and from 3.00 to 3.45, respectively. Reviser RL brings additional render success gains of 8.0 and 5.0 percentage points, reaching 96.6\% and 97.0\%, while maintaining comparable video quality (2.94 and 3.44). Trajectory supervision improves both render success and video quality, while RL primarily increases render success.

\keepspace{22\baselineskip}  
\begin{wraptable}{r}{0.42\linewidth}
    \vspace{-12pt}
    \centering
    \caption{Scene-module substitution in long-form systems. Both runs of a system share one lecture plan written by the untrained Qwen3.5-4B. \emph{Native} writes and repairs the scenes with the same untrained model, and \emph{+ Ours} generates them with our system.}
    \label{tab:exp-longform-transfer}
    \footnotesize
    \setlength{\tabcolsep}{4pt}
    \begin{tabular}{@{}lrr@{}}
        \toprule
        Metric (higher is better) & Native & + Ours \\
        \midrule
        \multicolumn{3}{@{}l}{\emph{Code2Video on MMMC}} \\
        Videos completed & 19/117 & \textbf{107/117} \\
        TeachQuiz gain ($\times$100) & $-$2.6 & \textbf{27.2} \\
        Aesthetics (0--100) & 35.5 & \textbf{58.7} \\
        \midrule
        \multicolumn{3}{@{}l}{\emph{TheoremExplainAgent on TEB}} \\
        Videos completed & 53/239 & \textbf{239/239} \\
        Visual score (1--5) & 2.35 & \textbf{3.16} \\
        \bottomrule
    \end{tabular}
    \vspace{-8pt}
\end{wraptable}

\paragraph{Transfer to long-form video generation.}
We use ManimCode as the scene generator in Code2Video and TheoremExplainAgent while keeping lecture plans and rendering budgets unchanged (Table~\ref{tab:exp-longform-transfer}).
Completed videos rise from 19/117 to 107/117 in Code2Video and from 53/239 to 239/239 in TheoremExplainAgent.
The evaluation also covers educational effectiveness and visual quality under the original benchmarks' dimensions. On MMMC, the TeachQuiz score, which measures quiz-based learning gain, rises from $-2.6$ to 27.2, and the aesthetics score rises from 35.5 to 58.7. On TEB, the visual score for visual relevance, element layout, and consistency rises from 2.35 to 3.16.

\begin{figure}[b]
    \centering
    \begin{minipage}[c]{0.49\linewidth}
        \centering
        \footnotesize
        \setlength{\tabcolsep}{3.2pt}
        \renewcommand{\arraystretch}{1.15}
        \begin{tabular}{@{}l r *{2}{>{\raggedleft\arraybackslash}p{2.6em}} r r r@{}}
            \toprule
            & & \multicolumn{2}{c}{ManimData-test} & \multicolumn{3}{c}{ManimBench} \\
            \cmidrule(lr){3-4}\cmidrule(l){5-7}
            Training data & Requests & RS$\uparrow$ & VQ$\uparrow$ & RS$\uparrow$ & VS$\uparrow$ & VQ$\uparrow$ \\
            \midrule
            HF Community & 13,768 & \textbf{76.0} & 2.76 & 58.0 & 50.4 & 2.81 \\
            JanusCode-800K & 6,289 & 23.2 & 2.38 & 43.0 & 40.4 & 3.26 \\
            ManimBench train & 315 & 2.0 & 2.10 & 22.0 & 20.9 & 2.99 \\
            \midrule
            ManimData (ours) & 24,034 & 64.4 & \textbf{2.82} & \textbf{81.0} & \textbf{69.7} & \textbf{3.29} \\
            \bottomrule
        \end{tabular}
    \end{minipage}\hfill
    \begin{minipage}[c]{0.48\linewidth}
        \centering
        \includegraphics[width=\linewidth]{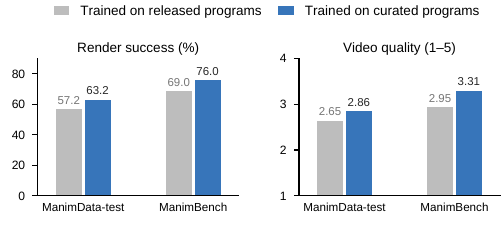}
    \end{minipage}
    \par\vspace{6pt}
    \captionsetup{position=below,skip=0pt}
    \begin{minipage}[t]{0.49\linewidth}
        \captionof{table}{Effect of training corpora. All models generate in a single pass. ManimData gives the highest RS and VS on ManimBench and the highest VQ on both benchmarks. Request counts exclude duplicates of evaluation requests (\supplref{app:sources}).}
        \label{tab:exp-data-ablation}
    \end{minipage}\hfill
    \begin{minipage}[t]{0.48\linewidth}
        \captionof{figure}{Effect of curation. On the same 8,416 community requests with identical inputs, a Qwen3.5-4B Writer fine-tuned on our curated programs (blue) scores higher in single-pass RS and VQ on both benchmarks than one fine-tuned on the released programs (gray).}
        \label{fig:exp-curation}
    \end{minipage}
\end{figure}

\subsection{Ablation Studies}
\label{sec:exp-ablations}

\paragraph{Training data and curation.}
Compared with existing Manim corpora (Table~\ref{tab:exp-data-ablation}), ManimData achieves the highest VQ on both benchmarks and the highest render success on ManimBench. We also train two Writers on the same 8,416 of the 8,497 community records, using either the original programs or their curated versions as supervision (Figure~\ref{fig:exp-curation}). The other 81 records are JanusCode-800K programs that define no \texttt{construct} method. On ManimData-test, curation raises render success from 57.2\% to 63.2\% and VQ from 2.65 to 2.86. On ManimBench, it raises render success from 69.0\% to 76.0\% and VQ from 2.95 to 3.31.

\subsection{Analysis}
\label{sec:exp-analysis}

\paragraph{Agreement with human preferences.}
To assess the reliability of our automatic quality judge, we conduct a human evaluation on 98 pairs of videos generated by ManimCode and GPT-4o or Qwen3.8-27B. A blinded annotator selects the video better suited for teaching or marks a tie (\supplref{app:human}).
Human preferences show stronger agreement with the judge when quality differences are larger: for VQ gaps of at least 0.5, the annotator agrees with the judge on 79\% of decided pairs, compared with only 29\% for gaps below 0.25. At the system level, human preferences align with the judge's rankings in three of four comparisons (Table~\ref{tab:app-human-pref}), favoring ManimCode over GPT-4o overall but Qwen3.8-27B on ManimData-test. The judge appears useful for assessing video quality, especially when the VQ gap is larger.

\begin{figure*}[t]
    \centering
    \includegraphics[width=\linewidth]{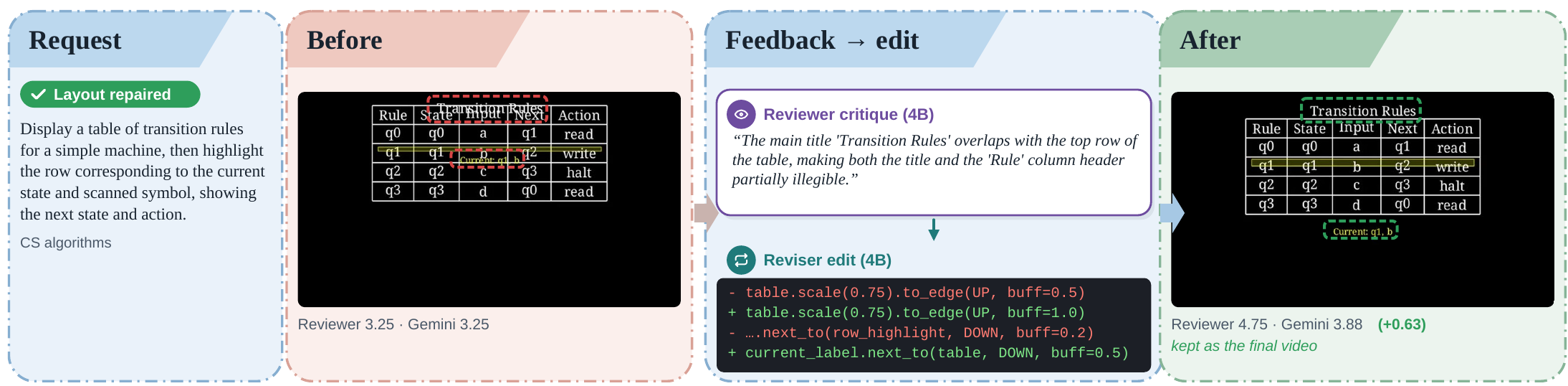}
    \caption{One revision step of our system on ManimData-test. Our Reviewer flags that the title overlaps the table header; the Reviser moves the table down and places the current-state label below it, and the revised video is kept as the final output. Dashed boxes mark the defect (red) and the repaired region (green). Under each frame are our Reviewer's mean score and Gemini 3.8 Flash score. More examples are in \supplref{app:qualitative}.}
    \label{fig:exp-qualitative}
\end{figure*}

\paragraph{Qualitative examples.}
In Figure~\ref{fig:exp-qualitative}, our Reviewer locates a layout defect in the rendered frames, the Reviser turns its critique into a small code edit, and the judge, which plays no part in the loop, also scores the revised video higher. Revisions can also fail. In one example of \supplref{app:qualitative}, the first of two consecutive revisions moves a list down to clear a progress bar but pushes it past the bottom edge, and the second moves it back up and restores the overlap. The Reviewer scores the second revision lower, so the loop keeps the first as its best candidate even though its list is still cut off. Keeping the best-scored candidate stops a regression from replacing a better version, but it cannot repair a defect that the Reviser leaves behind.

\section{Conclusion}
\label{sec:conclusion}

ManimData and ManimCode provide a data and training framework for feedback-driven educational animation generation. ManimData contains 24,034 render-verified, quality-filtered records that retain final programs alongside execution errors, visual critiques, and corrective revisions. ManimCode uses role-conditioned supervision and reinforcement learning on self-generated revision states to train a 4B model to write, review, and revise animation programs. It achieves 96.6\% and 97.0\% render success on ManimData-test and ManimBench, respectively, while maintaining competitive video quality against larger general-purpose models. When integrated into Code2Video and TheoremExplainAgent, ManimCode also improves long-form video completion and visual quality scores.

\clearpage
\bibliographystyle{plainnat}
\bibliography{ref}
\clearpage

\appendix

\renewcommand{\contentsname}{Appendix}
\tableofcontents
\addtocontents{toc}{\protect\setcounter{tocdepth}{2}}
\vspace{1.5em}

\section{Data}
\label{app:data}

\subsection{Request and Program Sources}
\label{app:sources}\label{app:funnel}

Wikipedia requests start from the November 1, 2023 dump of English Wikipedia. We keep the 664,747 articles with at least 1,000 average monthly page views and drop lists and timelines, which leaves 639,700. Qwen3.5-4B reads the introduction of each article and keeps 31,754 topics that describe a disciplinary concept, method, process, or structure. Qwen3.6-27B then keeps the 15,159 that could form a self-contained textbook section whose central ideas can be shown graphically. Qwen3.8-27B finds directly drawable content in 12,516 of these topics and writes 24,770 requests for them. A further drawability check on biology, anatomy, and medicine removes the requests that depend on realistic illustration, which leaves 22,058 requests on 11,150 topics.

The seven Hugging Face community datasets are \emph{manim-code-maths}, \emph{cleaned-manim-dataset}, \emph{3blue1brown-manim}, \emph{manim-code-dataset}, \emph{manim-auto-dataset}, \emph{manim-finetune-data}, and \emph{autoresearch-manim}. After merging and deduplication they hold 20,976 examples, of which 14,269 render as released. We keep these and hold out 500 of them, from which the 281 community requests of ManimData-test are drawn; its 219 Wikipedia requests come from topics that never enter training. JanusCode-800K contributes its 8,316 Manim examples. From ManimBench v1 we take the 316-example training split and exclude the 100-example test split. For each community source, we strip Markdown code fences and trailing end-of-generation markers and remove exact duplicate request-program pairs, which leaves 13,768, 6,365, and 316 inputs, respectively. The corpus rows of Table~\ref{tab:exp-data-ablation} train on each source as released, after removing requests that duplicate an evaluation request under text normalization (76 from JanusCode-800K and 1 from ManimBench), which leaves 13,768, 6,289, and 315 training examples.

Table~\ref{tab:manimdata-funnel} follows each source through construction.

\begin{table}[!htb]
    \centering
    \begin{minipage}[t]{0.555\linewidth}
        \centering
        \caption{\textbf{ManimData construction by source.} \emph{1st renders}: share of inputs whose first version (the released code for community sources) renders a video. \emph{Admitted}: inputs whose best version has a mean score $\geq 3.75$. \emph{Final}: admitted records left after removing overlap with the evaluation sets. \emph{Unchanged}: share of final records whose first version is the final program. $^{\dagger}$Most ManimBench inputs whose first version renders no video are static scenes that save only an image, which ManimBench's own protocol counts as rendered.}
        \label{tab:manimdata-funnel}
        \scriptsize
        \setlength{\tabcolsep}{3pt}
        \begin{tabular}{@{}lrrrrr@{}}
            \toprule
            Source & Inputs & 1st renders & Admitted & Final & Unchanged \\
            \midrule
            Wikipedia & 22,058 & 56.5\% & 15,908 & 15,537 & 13.1\% \\
            HF Community & 13,768 & 100.0\% & 4,284 & 4,284 & 22.6\% \\
            JanusCode-800K & 6,365 & 56.4\% & 4,018 & 3,944 & 4.3\% \\
            ManimBench train & 316 & 41.1\%$^{\dagger}$ & 269 & 269 & 15.2\% \\
            \midrule
            Total & 42,507 & 70.5\% & 24,479 & 24,034 & 13.4\% \\
            \bottomrule
        \end{tabular}
    \end{minipage}\hfill
    \begin{minipage}[t]{0.405\linewidth}
        \centering
        \caption{\textbf{Composition of the API knowledge base.} Number of API cards by source and entry type, after deduplication.}
        \label{tab:app-kb}
        \scriptsize
        \setlength{\tabcolsep}{3pt}
        \begin{tabular}{@{}llr@{}}
            \toprule
            Source & Entry type & Cards \\
            \midrule
            \multirow{5}{*}{ManimCE docs} & Method or property & 822 \\
             & Class & 358 \\
             & Function or constant & 218 \\
             & Guide section & 115 \\
             & Module overview & 56 \\
            \midrule
            \multirow{3}{*}{Eight plugins} & Object card & 203 \\
             & Guide section & 132 \\
             & Capability recipe & 21 \\
            \midrule
            \multirow{2}{*}{manim\_skill} & Best-practice rule & 233 \\
             & Example scene & 61 \\
            \midrule
            Total & & 2,219 \\
            \bottomrule
        \end{tabular}
    \end{minipage}
\end{table}

\subsection{API Knowledge Base}
\label{app:kb}

The knowledge base covers ManimCE and eight plugins (manim-physics, manim-chemistry, manim-ml, manim-dsa, manim-eng, manim-circuit, manim-automata, and manim-code-blocks), each verified to render with ManimCE v0.19.1. Core cards come from the ManimCE v0.19.0 documentation (541 pages, 379 of them API reference), which we use with the v0.19.1 renderer. We parse its Sphinx markup into one card per class, method, function, module overview, and guide section, with code examples kept verbatim. For each plugin object, Qwen3.8-27B reads the source code, docstring, and usage snippets, labels the object as user-facing or internal, and writes its purpose, parameters with example values, required scene class, call order, caveats, and a minimal example scene. We render each example scene offline and retry up to three times with the traceback as feedback. Of the 336 public objects, 203 are user-facing, and 175 of these have a render-verified example. The 21 capability recipes map topic keywords (e.g., \emph{electric field}, \emph{Coulomb}) to a plugin, its import line, and a verified code fragment. Table~\ref{tab:app-kb} counts the cards by source and entry type after deduplication.

\subsection{Construction Loop Settings}
\label{app:loop}

Table~\ref{tab:app-loop} lists the construction-loop settings (Sec.~\ref{sec:manimdata-curation}). At inference, our model plays every role in the render--review--revise loop and follows the same settings except for Admission (Sec.~\ref{sec:method-rl}).

\begin{table}[!htb]
    \centering
    \caption{\textbf{Construction loop settings.} Qwen3.8-27B plays every role. The Writer row also applies to the Reviser, which rewrites the full program from the error or the critique. \emph{Acceptance} decides when a candidate replaces the best candidate $c^*$, and \emph{Admission} decides whether a record enters ManimData.}
    \label{tab:app-loop}
    \footnotesize
    \begin{tabularx}{\linewidth}{@{}l >{\raggedright\arraybackslash}X@{}}
        \toprule
        Setting & Value \\
        \midrule
        Renderer & ManimCE v0.19.1 with eight domain plugins, 480p \\
        Budget & 10 renders per input \\
        Writer & Qwen3.8-27B; writes the full program from retrieved cards and feedback \\
        Retrieval & 2--4 queries per step; Qwen3-Embedding-8B search, Qwen3-Reranker-8B reranking; top 10 cards \\
        Layout checks & Objects outside the frame, overlapping text, labels too small to read, at animation pause points \\
        Reviewer & Qwen3.8-27B; request, code, and 8 uniformly sampled frames; four dimensions scored 1--5 \\
        Acceptance & Mean score increases; request alignment and content completeness drop by at most 0.5 \\
        Revision base & Latest candidate, unless it lost more than 0.5 in request alignment or content completeness relative to the best \\
        Early stop & Mean $> 4.0$ and every dimension $> 3.5$, confirmed by a second review on a shifted set of 8 frames \\
        Admission & Best candidate's mean score $\geq 3.75$ \\
        \bottomrule
    \end{tabularx}
\end{table}

\clearpage
\section{Method}
\label{app:method}

\subsection{Role-Conditioned Supervision}
\label{app:supervision}

We extract five types of examples from the 24,034 ManimData trajectories (Table~\ref{tab:method-supervision}). A writing example pairs the request and its retrieved API cards with the final program of the record. Community records lack a Writer step because their first version is released code. To reconstruct the Writer input, Qwen3.8-27B writes retrieval queries for the request, and the top 10 retrieved cards form the context. An execution-repair example keeps a revision that turns a crashing program into one that renders. A quality-revision example keeps a revision whose mean Reviewer score is higher than that of the program it revised. We subsample reviewing examples from the recorded reviews, with at most two per community record. Retrieval-query examples pair Qwen3.8-27B's queries with the request and any error or critique.

\begin{table*}[!htb]
    \centering
    \caption{\textbf{Role-conditioned supervision.} Input context, target output, and number of examples for each supervision type extracted from the 24,034 ManimData trajectories.}
    \label{tab:method-supervision}
    \footnotesize
    \setlength{\tabcolsep}{5pt}
    \renewcommand{\arraystretch}{1.15}
    \begin{tabularx}{\textwidth}{@{}l >{\raggedright\arraybackslash}X >{\raggedright\arraybackslash}X r@{}}
        \toprule
        Supervision & Input context & Target output & Examples \\
        \midrule
        Writing & Request; retrieved API cards & Final program selected during construction & 24,034 \\
        Execution repair & Request; failed code; execution error; API cards & Renderable repaired program & 14,587 \\
        Quality revision & Request; code; scores and critique; layout diagnostics; API cards & Higher-rated revised program & 32,683 \\
        Reviewing & Request; code; sampled video frames & Scores on the four review dimensions and a critique & 46,965 \\
        Retrieval queries & Request; error or critique when present & Two to four API search queries & 187,183 \\
        \midrule
        Total & & & 305,452 \\
        \bottomrule
    \end{tabularx}
\end{table*}

\subsection{Training Details}
\label{app:training}

Table~\ref{tab:app-training} gives the hyperparameters of both training stages: supervised fine-tuning with the loss of Eq.~\eqref{eq:method-sft} and revision RL with RLOO (Sec.~\ref{sec:method-rl}). Supervised fine-tuning takes about 32 hours on four NVIDIA L40S GPUs. Reinforcement learning runs on the same four GPUs and takes about 8 minutes per step; the reward model (Qwen3.8-27B) is served on a separate machine. Thinking mode is disabled throughout.

\begin{table}[!htb]
    \centering
    \caption{\textbf{Training hyperparameters.} Both stages train LoRA adapters~\cite{hu2022lora}; the RL adapter starts from the merged supervised model.}
    \label{tab:app-training}
    \footnotesize
    \setlength{\tabcolsep}{4pt}
    \begin{tabular}{@{}ll@{\hspace{2.5em}}ll@{}}
        \toprule
        \multicolumn{2}{@{}l}{\emph{Supervised fine-tuning}} & \multicolumn{2}{l}{\emph{Revision RL}} \\
        \midrule
        Base model & Qwen3.5-4B & Initialization & supervised model \\
        LoRA & rank 32, $\alpha=64$, dropout 0.05 & LoRA & rank 32, $\alpha=64$ \\
        Target modules & all linear layers & Advantage & leave-one-out (RLOO) \\
        Learning rate & $1\times10^{-4}$, cosine & Learning rate & $1\times10^{-4}$ \\
        Warm-up & 10 steps & Batch & 16 states $\times$ $G=8$ samples \\
        Batch size & 16 sequences & Policy updates & 2 per step (8 states each) \\
        Epochs & 1 & Clip range & $[1-0.2,\ 1+0.28]$ \\
        Max length & 10,240 tokens & KL to supervised policy & 0.01 (low-variance estimator) \\
        Precision & bf16 & Entropy bonus & 0 \\
        & & Sampling & temperature 1.0, top-$p$ 0.95 \\
        & & Max response & 4,096 tokens \\
        & & Gradient clipping & 1.0 \\
        & & Steps & 40 (checkpoint at step 20 used) \\
        \bottomrule
    \end{tabular}
\end{table}

\clearpage
\section{Evaluation}
\label{app:eval}

\subsection{Baselines and Model Versions}
\label{app:eval-baselines}

Single-pass baselines decode at temperature 0.4 and top-$p$ 0.95 with at most 8,192 new tokens; thinking is disabled for models that support it. On ManimData-test they receive our Writer's system prompt without retrieved API cards; on ManimBench they receive the benchmark's own prompt. GPT-4o is the snapshot \texttt{gpt-4o-2024-11-20}. GLM-5.3 Flash (\texttt{z-ai/glm-5.3-flash}) and DeepSeek-V4 Flash (\texttt{deepseek/deepseek-v4-flash}) are queried through OpenRouter in November 2026 with the same sampling settings, pinned to the official Z.AI endpoint (fp8) for GLM-5.3 Flash and to DeepInfra (fp8) for DeepSeek-V4 Flash, with Fireworks as the only allowed alternative; Fireworks served 15 of the 600 DeepSeek-V4 Flash requests. GLM-5.3 Flash runs at minimal reasoning effort (149 and 11 reasoning tokens per request on average on ManimData-test and ManimBench), and DeepSeek-V4 Flash with reasoning off. JanusCoder-8B is the released \texttt{internlm/JanusCoder-8B}. For ManimTrainer we use its authors' released LoRA adapter for \texttt{ByteDance-Seed/Seed-Coder-8B-Instruct}, trained with SFT and GRPO, and run it with their chat template. In the RITL-DOC row, ManimTrainer uses its authors' inference-time repair loop with documentation retrieval, which allows at most three feedback rounds and at most four renders. The row for the strongest ManimTrainer configuration uses Qwen3-Coder-30B-A3B, trained with GRPO under RITL-DOC with three rounds, and reports the ManimBench render success and visual similarity from Table~3 of its paper. That model is not released, and the authors used ManimCE 0.19.0; we use 0.19.1.

\subsection{Judge}
\label{app:eval-judge}

The judge is Gemini 3.8 Flash (API model \texttt{models/gemini-3.8-flash}). It sees the request and the rendered video at 2 frames per second, but not the code. One call assigns 1--5 scores for request alignment, content completeness, and semantic correctness; their mean is $Q_{\mathrm{cont}}$. A dedicated presentation prompt scores visual clarity and animation quality in two calls; $Q_{\mathrm{pres}}$ is the mean of the calls' averages. VQ is the mean of $Q_{\mathrm{cont}}$ and $Q_{\mathrm{pres}}$. The scene-level scores in each table come from one batch run in October 2026, except the GLM-5.3 Flash and DeepSeek-V4 Flash rows of Table~\ref{tab:exp-scene-results}, which were scored on November 1, 2026. For that later batch, we re-scored 30 ManimData-test and 30 ManimBench anchor videos from October; their mean VQ changed by $+0.03$ and $-0.02$.

\subsection{Long-Form Evaluation}
\label{app:eval-longform}

For each long-form system, both runs use a lecture plan written by the untrained Qwen3.5-4B in that system's native configuration; they differ only in the scene generator. The native run uses that model to write and repair each scene, while our run uses ManimCode. Each run gets at most 10 renders per section or scene, and sections that never render are skipped when the video is merged. We replace each benchmark's original judge with Gemini 3.8 Flash, so absolute scores are not comparable with published numbers.

For MMMC we report the number of completed videos, the TeachQuiz learning gain ($\times$100) over all 117 topics, and the aesthetics score (0--100) over the produced videos; a topic without a video counts as zero TeachQuiz gain. The 4B planner produces no valid plan for 10 of the 117 topics, so our 107 videos cover every topic that has a plan. TheoremExplainBench (TEB) has 239 distinct theorems; its list of 240 contains \emph{The Ideal Gas Law} twice. Its visual score is the geometric mean (1--5) of three visual dimensions (visual relevance, element layout, and visual consistency) over the produced videos, judged with the audio removed; an all-black video scores 1 on the two frame-based dimensions. We omit the two narration-based dimensions of TEB because they score the shared plan. For 10 theorems, the native run never saved the implementation plans of some scenes (23 of the 1,371), and our run re-planned these scenes with the same untrained 4B model.

\subsection{Human Evaluation Protocol}
\label{app:human}

We sampled at random (with a fixed seed) 98 requests that both our system and its opponent rendered: 48 against GPT-4o (31 from ManimData-test and 17 from ManimBench) and 50 against Qwen3.8-27B (36 and 14); no request was used for both opponents. The videos are the files scored in Table~\ref{tab:exp-scene-results}, so each human choice can be compared with the judge's scores for the same two videos.

Each page of the annotation interface shows the full request and the two silent videos side by side, labeled A and B in random order, without system names or scores. Both videos loop, and an answer is accepted only after each video has played for at least 90\% of its length; dragging the scrubber to the end does not count. The annotator chooses the video they would rather show to students to explain the request (A, about the same, or B). They weigh whether the content is correct and covers the request, and whether everything is readable, inside the frame, and free of overlaps; the instructions state that a longer video is not better by itself. Pairs are shown in random order, and two of them reappear later as a consistency check; the repeats are excluded from all counts.

One member of our team, who did not know which system produced which video, judged all 98 pairs, taking a median of 24 seconds per pair. Table~\ref{tab:app-human-gap} splits the choices by the judge's VQ gap, and Table~\ref{tab:app-human-pref} compares the human and judge verdicts per opponent and benchmark. Among pairs in which both the annotator and the judge picked a side, the annotator agreed with the judge's content ordering in 69\% and with its presentation ordering in 65\%. Neither the annotator nor the judge showed a detectable preference for longer videos. Regressing the annotator's verdict (our video preferred $=1$, about the same $=0.5$, the opponent's preferred $=0$) on the $\log_2$ ratio of the two video lengths, with opponent-by-benchmark fixed effects, gives a slope of 0.03 per doubling (95\% CI $-0.09$ to $0.13$); the same fit on the judge's verdicts gives 0.05 ($-0.08$ to $0.16$). All intervals are percentile bootstrap intervals over pairs.

\begin{table*}[!htb]
    \centering
    \small
    \caption{\textbf{Human choices by VQ gap.} The gap is the absolute difference between the VQ scores that the judge (Gemini) gave the two videos of a pair (98 pairs, one annotator). Among pairs in which both the annotator and the judge chose a side, agreement is the share in which the annotator picked the video the judge scored higher (95\% bootstrap interval in brackets). In the 14 pairs with equal judge scores, the annotator answered about the same in 6 and chose a side in 8; in the All row, those 8 count only in the Pairs column.}
    \label{tab:app-human-gap}
    \begin{tabular}{lrrrrc}
        \toprule
        Gemini VQ gap & Pairs & Gemini's higher & About the same & Gemini's lower & Agreement (\%) \\
        \midrule
        0 (equal scores) & 14 & -- & 6 & -- & -- \\
        $(0, 0.25)$ & 19 & 4 & 5 & 10 & 29 [7, 50] \\
        $[0.25, 0.5)$ & 26 & 15 & 5 & 6 & 71 [52, 90] \\
        $[0.5, 1)$ & 31 & 22 & 3 & 6 & 79 [61, 93] \\
        $\ge 1$ & 8 & 5 & 2 & 1 & 83 [50, 100] \\
        \midrule
        All & 98 & 46 & 21 & 23 & 67 [55, 78] \\
        \bottomrule
    \end{tabular}
\end{table*}

\begin{table*}[!htb]
    \centering
    \small
    \caption{\textbf{Human and judge (Gemini) verdicts on the same pairs.} Wins, ties, and losses of our system against each opponent; for the judge, a tie means equal VQ scores. The human win rate counts a tie as half (95\% bootstrap interval in brackets).}
    \label{tab:app-human-pref}
    \begin{tabular}{llrccc}
        \toprule
        Opponent & Benchmark & Pairs & Human W / T / L & Human win rate & Gemini W / T / L \\
        \midrule
        GPT-4o & ManimData-test & 31 & 18 / 4 / 9 & 0.65 [0.48, 0.81] & 17 / 7 / 7 \\
        GPT-4o & ManimBench & 17 & 7 / 7 / 3 & 0.62 [0.44, 0.79] & 9 / 3 / 5 \\
        Qwen3.8-27B & ManimData-test & 36 & 11 / 2 / 23 & 0.33 [0.19, 0.49] & 14 / 3 / 19 \\
        Qwen3.8-27B & ManimBench & 14 & 5 / 8 / 1 & 0.64 [0.50, 0.79] & 6 / 1 / 7 \\
        \bottomrule
    \end{tabular}
\end{table*}

\keepspace{14\baselineskip}  
\subsection{Qualitative Revision Examples}
\label{app:qualitative}

\begin{figure*}[!b]
    \centering
    \includegraphics[width=0.95\linewidth]{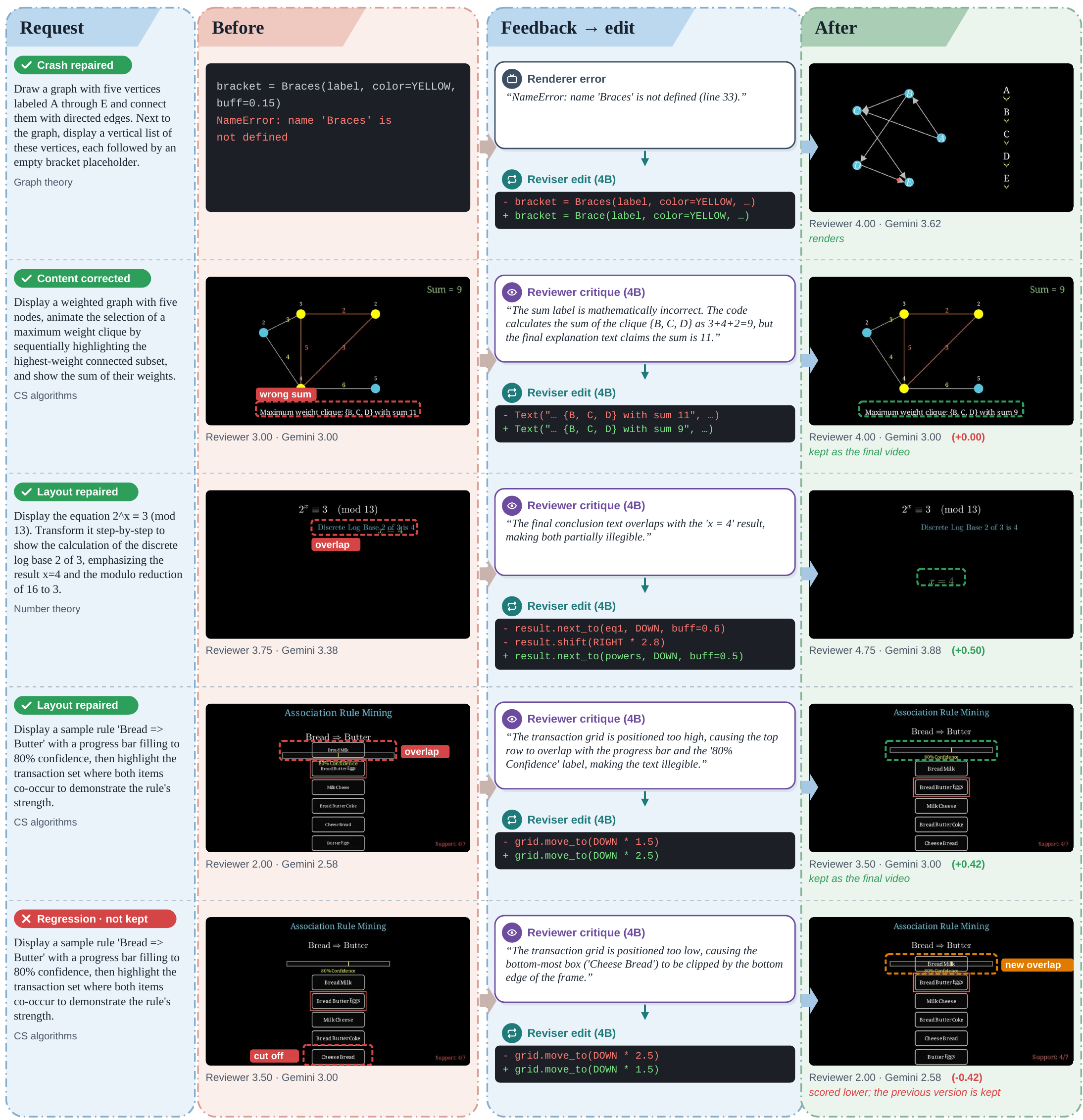}
    \caption{\textbf{More revision steps of our system on ManimData-test.} Each row is one revision step; the best-candidate rule decides which version the loop returns. From the top: a crash repaired from the renderer's error message; a wrong sum corrected in the explanation text; an overlap removed by moving the result below the derivation; and two consecutive revisions of the same request. In the last two rows, moving the transaction list down clears the progress bar but pushes the list past the bottom edge, and moving it back up restores the overlap; our Reviewer scores the second revision lower, so the loop keeps the first. Under each frame are our Reviewer's mean score and the score that the judge, Gemini 3.8 Flash, gives the re-rendered video (1--5).}
    \label{fig:app-qualitative}
\end{figure*}

Figure~\ref{fig:app-qualitative} shows five more revision steps from our system on ManimData-test, in the same format as Figure~\ref{fig:exp-qualitative}. We automatically drew candidate steps from the evaluation run by edit type: crash repair, layout repair, content correction, and revisions that our Reviewer scored lower. We re-rendered both versions of each candidate for the judge, then kept steps whose defect was visible in a sampled frame. When both versions had a score, the judge's score moved in the same direction as our Reviewer's, except for the content correction, which the judge scored 3.00 before and after.

\end{document}